# 3D Quantum Cuts for Automatic Segmentation of Porous Media in Tomography Images

Junaid Malik[a], Serkan Kiranyaz[a], Riyadh I. Al-Raoush[a], Olivier Monga[b], Patricia Garnier[c], Sebti Foufou[d], Abdelaziz Bouras[a], Alexandros Iosifidis[e], Moncef Gabbouj[f], Philippe C. Baveye[g]

[a], Qatar University, Doha, 2713, Qatar
[b], Institute of Research for Development, Marseille, France
[c], French National Institute for Agricultural Research, INRA, France
[d], University of Burgundy, France
[e], Aarhus University, Aarhus, Denmark
[f], Tampere University of Technology, Tampere, Finland
[g], AgroParis-Tech, France

**Abstract**— Binary segmentation of volumetric images of porous media is a crucial step towards gaining a deeper understanding of the factors governing biogeochemical processes at minute scales. Contemporary work primarily revolves around primitive techniques based on global or local adaptive thresholding that have known common drawbacks in image segmentation. Moreover, absence of a unified benchmark prohibits quantitative evaluation, which further clouds the impact of existing methodologies. In this study, we tackle the issue on both fronts. Firstly, by drawing parallels with natural image segmentation, we propose a novel, and automatic segmentation technique, 3D Quantum Cuts (QCuts-3D) grounded on a state-of-the-art spectral clustering technique. Secondly, we curate and present a publicly available dataset of 68 multiphase volumetric images of porous media with diverse solid geometries, along with voxel-wise ground truth annotations for each constituting phase. We provide comparative evaluations between QCuts-3D and the current state-of-the-art over this dataset across a variety of evaluation metrics. The proposed systematic approach achieves a 26% increase in AUROC while achieving a substantial reduction of the computational complexity of the state-of-the-art competitors. Moreover, statistical analysis reveals that the proposed method exhibits significant robustness against the compositional variations of porous media.

## 1 Introduction

Rapid advances in micro-CT technology and easier access to scanning hardware have made X-ray computed tomography imaging quite popular among soil scientists [1,2]. This technique enables and facilitates the visualization, as well as the subsequent quantitative analysis, of the characteristics of porous media in a non-invasive manner at micrometric scales. Data obtained from tomography is generally in the form of grayscale 3D volumetric images where the intensity at each spatial location in the digital representation is proportional to the X-ray attenuation properties of the material present at the corresponding location in the scanned sample. These high-resolution volumetric images enable researchers to model, predict and better understand the biogeochemical processes occurring within the porous media at fine scales [1,3,4]. An essential step in this quest is the accurate segmentation, which aims at identifying particular populations of voxels, associated with distinct phases or constituents. The most common approach in this regard is the binarization (or binary segmentation), which consists of distinguishing two phases corresponding to pore space (void) and solids.

- J. Malik (Email: hafiz.malik@qu.edu.qa), S. Kiranyaz (Email: mkiranyaz@qu.edu.qa), R.I. Al-Raoush (Email: riyadh@qu.edu.qa) and Abdelaziz Bouras (Email: abdelziz.bouras@qu.edu.qa) are with Qatar University, P.O. Box 2713, Doha, Qatar.
- O. Monga (Email: olivier.monga@ird.fr) is with the Institute of Research for Development, Marseille, France.
- P. Garnier (Email: patricia.garnier@inra.fr) is with French National Institute for Agricultural Research, INRA.
- Sebti Foufou (Email: sf146@nyu.edu) is with the University of Burgundy, France.
- A. Iosifidis (Email: alexandros.iosifidis@eng.au.dk) is with Aarhus University, Aarhus, Denmark.
- M. Gabbouj (Email: moncef.gabbouj@tut.fi) is with Tampere University of Technology, Tampere, Finland.
- P. Baveye (Email: philippe.baveye@agroparistech.fr) is with AgroParisTech, France.

Over the years, a substantial amount of research has been conducted towards testing existing image processing methods as well as devising novel schemes to tackle this binary segmentation problem [5–7]. Prevalent methods essentially employ a thresholding operation to recognize voxels belonging to pore space. These methods are commonly categorized as either global, when they employ a single thresholding value to categorize voxels, and local, when they aim to label individual voxels based on the intensity information in their local neighborhood. Comprehensive visual and quantitative comparisons among 14 different segmentation methods were carried out earlier in [5]. Among the global-thresholding methods, clustering-based method of [8] and the iterative technique proposed in [9] showed promising results. However, such methods have been proven to be less robust to noisy artifacts in the images such as global intensity variation due to poor contrast and partial volume effect, and thus, they fail to produce consistently stable results. Local methods tend to produce better overall segmentation outcomes as they account for the local spatial image information. Noteworthy among them are the indicator kriging based methods [10,11], region growing based method of [12], and an adaptive variation of [8] for a localized application [13]. Indicator kriging [10] is one of the most widely used method among them. However, it is sensitive to the initial pre-classification step that requires a skilled operator to define lower and upper thresholds for pore and solid voxels respectively. Efforts have been made, such as in [12], to infer the initial thresholds from low-level image properties such as edge information. In spite of this, as noted by [13], the efficacy of these methods still depends on factors such as choice of pre-filtering and tuning of design parameters, which inevitably leads to subjectivity in their use. The method proposed by [13] is the only fully automatic segmentation regime proposed thus far, but it has the added complexity of exhaustively searching for optimal sets of parameters. Collectively, all the fore-mentioned techniques assume a bimodal behavior of the histogram, either globally or within local windows. This is a common drawback among all thresholding-based segmentation methods as they are inherently reliant on the histogram shape and fail to provide adequate results when dealing with complex distributions. This is primarily why thresholding is rarely used for 2D segmentation tasks for natural images. This is echoed by the findings of numerous studies which report a lack of efficient, unsupervised segmentation regimes that are tailored specifically for volumetric images of porous media [5–7,14].

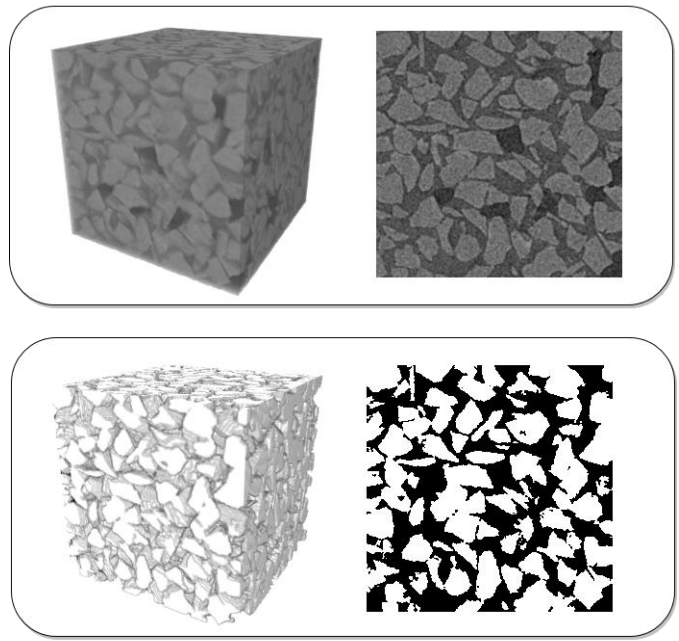

Figure 1 3D visualization (left) and 2D slice (right) of a test volumetric image (top) and its segmented output (bottom).

Salient-object detection is a major area of interest in computer vision research that deals with identifying visually unique and prominent regions in natural images [15]. Prevailing methods exploit a variety of local and global cues such as edge information, color contrast and spatial location to classify image pixels as belonging to either the foreground region (salient-object) or the background. One can observe that this task is similar to the binary segmentation of porous media representations in the key sense that both are concerned with separating out a single region of interest (salient-object/solids) from the backdrop (background/pore-space) in a one-vs-rest fashion. Moreover, the region of interest in both cases is texturally homogenous and unique from the immediate spatial surroundings. Based on this, we argue that salient object detection methods are potentially more suitable for application to porous media images as opposed to classical thresholding techniques. It is worth noting that, quite unlike the segmentation of porous media images, the field of salient object detection is ripe with numerous novel contributions, both in terms of diversity of segmentation techniques and number of publicly available benchmark datasets. In an extensive exploratory work undertaken by [16], 29 methods were evaluated on 7 benchmark datasets across multiple evaluation metrics. Among the tested methods, Quantum Cuts (QCuts) [17–19], a recent segmentation technique based on Quantum Mechanical principles, consistently stood out as the best performing method. Albeit unsupervised, QCuts provided results *on par* with the supervised techniques, and even outperformed them over some datasets. In light of these findings, we hypothesize that an automatic, unsupervised and class-independent salient object detection method like QCuts lends itself to application for segmenting volumetric tomography images of porous media.

Furthermore, in addition to the lack of efficient segmentation systems, another critical factor hindering progress in the field of porous media images' segmentation is the absence of a benchmark dataset with annotated ground truth that can be used to qualitatively and quantitatively evaluate 3D porous segmentation methods. Curation of such datasets is made harder because of the fact that, for the case of naturally occurring porous media such as real soils, obtaining a voxel-wise ground truth is not possible [7,13]. Morphological characteristics such as the porosity (percentage of space occupied by pores) of a real sample and that calculated from a segmented image are often compared to gauge the segmentation accuracy. However, such measures are not adequate as the total porosity of a soil sample is affected by pores that might not be visible at the studied X-ray resolution [2,7]. As a workaround, researchers have resorted either to generating synthetic soil images of known porosity by working backward from binary images [7], or to using artificial systems whose porosities can be manually controlled [5]. Moreover, the porosity measure itself is not descriptive enough as it only takes into account the overall percentage of pores, not their spatial location. This absence of benchmark datasets renders the quantifiable comparison among the segmentation results impossible.

Collectively, the evidence presented above highlights key issues concerning both the outdation of segmentation techniques and the uncertainity regarding their performance that arises from the absence of annotated datasets. In this study we address these two issues and make the following novel contributions:

i) QCuts-3D: an extension by 3D reformulation of the state-of-the-art 2D salient object detection method of QCuts, tailored exclusively for segmenting volumetric tomography images of porous media,
ii) SIMUPOR dataset: a collection of high-resolution multiphase volumetric images along with expert annotated voxel-wise ground truth for each phase.

We test our method against the current state-of-the-art method of [13] on this new dataset and calculate a variety of performance metrics that are commonly used in segmentation problems.

## 2 Prior Work

Several earlier exploratory studies have evaluated the performance of a variety of image segmentation methods on images of porous media. However, very few of these methods have been designed specifically for the segmentation of volumetric representations of porous materials. Many methods employ a 2D "slice-by-slice" approach, which is known to have several drawbacks [6,20] and is also prone to directional bias [21]. Therefore, for the sake of brevity, in this study we restrict our discussion only to those methods that truly operate in 3D volumetric space and have been proposed exclusively for the task of segmentation of porous media representations. A detailed account of the methods omitted from this discussion can be found in [5–7]. The latter part of this section provides a brief overview of unsupervised salient object detection along with the rationale behind adopting Quantum Cuts.

In one of the earliest efforts related to soil segmentation, [22] proposed a global histogram-based approach. Assuming a bimodal distribution of the grayscale intensities, two threshold values are calculated; $T_{min}$ and $T_{max}$. All voxels with intensities below $T_{min}$ are labeled as belonging to pore space, whereas the ones above $T_{max}$ are labeled as solids. The identified pore voxels then act as an initial seed for growing the pore space region whereby all neighboring voxels with intensity less than $T_{max}$ are iteratively added to the pore space region. Later on, the authors of [12] proposed an alternative approach to calculating the initial thresholds that are less affected by the shape of the histogram. As a pre-processing step, smoothing based on pseudomedian filter is applied, aimed at suppressing noise while preserving the edges. Binary edge maps are then calculated, which identify the regions with high intensity gradients by applying Sobel [23] and Laplace-based edge-detection methods. The two initial thresholds, $T_{min}$ and $T_{max}$ are then calculated from the histograms of each of the two gradient masks and averaged to obtain two final values. The rest of the segmentation ensues as described in [22]. Evaluation of the proposed method was performed using synthetic test images with predefined porosities. Images were also degraded with various degrees of noise to make the segmentation more challenging. The performance was measured based on the discrepancy in the porosity values.

Later, Sheppard et al. in [24], proposed a multi-stage pipeline consisting of a two-step preprocessing followed by segmentation. The given volumetric image is first smoothed using anisotropic filtering for noise reduction. In order to recover high-frequency information lost as a result of smoothing, unsharp masking is applied for enhancement of edges. Finally, for segmentation, a combination of watershed transform [25] and a modified version of seeded region growing based on active contours [26] was employed. Specifically, a modified energy function is proposed based on image gradients and intensity levels. For tracking the evolution of the segmentation boundary, fast marching algorithm [27] is used and the stopping criterion is based on the common boundary shared between the two interfaces.

The authors of [11] proposed a variation on top of the widely used method of [10]. Recognizing the computational cost and subjectivity associated with applying indicator kriging using a window size of constant radius, the authors proposed to adapt the parameter based on local image conditions. Given a volumetric image, the voxels are first partly preclassified using a pair of thresholds as in [10]. Afterward, to model the spatial variance, an empirical semivariogram is obtained for the whole image and a theoretical model is estimated by fitting it to the empirical data. Labeling for the unclassified voxels is then obtained by calculating class probabilities using this model (kriging system). In the original framework of [10], the problem of inconclusive class probabilities is addressed by multiple passes over the image with the same window size, rendering the process computationally expensive. To remedy this, the authors of [11] propose adapting the window size progressively until a satisfactory labeling for the voxels inside the window is obtained. A majority filtering operation is also applied after the pre-classification and kriging steps, as recommended in [28]. The method was tested on 5 soil images and was shown to achieve similar results to [10] with significantly less computational cost.

In [13], the global method of [8] is extended and applied in a localized manner. A given volumetric image is first decomposed into non-overlapping cubes of fixed sizes. For each cube, a threshold value $T_{solid}$ is calculated based on the profiles of the intra-class variance and the solid and pore phase variance functions. Afterwards, an interpolation operation is applied to smooth the thresholding surface, which is finally used to obtain the desired binary segmentation. A majority filtering operation with a threshold of 60% is applied as a postprocessing step to remove isolated solid/pore voxels. As is the case with most of the local methods, the critical design parameter is the window size used for decomposing the original image. To make the choice operator-independent, the authors propose an automatic approach where a number of window sizes are first used and an optimal value is then chosen in an unsupervised fashion based on the proposed selection criterion. In the evaluation performed over a variety of synthetic soil images, the proposed method is shown to achieve a more accurate estimate of the porosity values and a lower misclassification error, compared to competing methods.

An overview of existing methodologies reveals an obvious need for more robust and efficient unsupervised segmentation regimes. Performance of indicator kriging based methods [10,11] relies heavily on the pair of thresholds chosen for pre-classification. Despite efforts made to automate this step, as noted by [11], expert intervention is still required to get reasonable and stable outcomes. For the case of pipelines involving preprocessing [12,22,24], there is a significant degree of operator subjectivity related to the choice of filtering operation and tuning of the parameters involved. Region growing-based methods, such as the active contour-based method of [24], are inherently sensitive to the choice of initialization. The method of [13] is the only fully automatic solution that does not involve manual decisions regarding the tuning of the parameters. This is achieved by exhaustively searching the parameter space for an optimum value that involves multiple passes over the image, which is not desirable. In addition to the size of the local window, the shape is also quite critical as cubic or spherical windows as used in [11] and [13] cannot guarantee that the contours of the solid grains will be preserved. Finally, the most crucial drawback is that all these threshold-based segmentation methods assume that the volumetric image has two peaks in the gray-scale histogram or equivalently, there are two major gray scale values, one for pores and the other for the solid. For those images where the solid intensity varies and/or the pores are filled with different media (gas, saline water or oil), this will obviously not be the case and hence pores tend not to be segmented properly. In general, any thresholding-based operation, global or local, is histogram dependent and entirely overlooks spatial information. This is one of the key reason behind the obsolence of thresholding-based techniques since the earlier image segmentation efforts.

One of the most widely studied natural 2D image segmentation problems over the recent past is the task of salient object detection, which deals with identifying a single region of interest that is visually appealing and salient from the spatial surroundings. Quite similar to binary segmentation of porous media images, salient object detection aims at a one-vs-rest categorization of the image elements into the foreground (region-of-interest) and the background. A wide variety of salient object segmentation literature exists, along with numerous labelled datasets for benchmarking. In a thorough quantitive evaluation of 29 salient object detection over 7 different datasets, the methods of DRFI [29], RBD [30], ST [31], DSR [32], MC [33] and QCUT [19], came out as the top performing methods; [19] being the best among unsupervised methods. All these works exploit a background prior based on the fact that the boundary of a natural 2D image is more likely to belong to the background. In [32], superpixels residing on the edges of the image are used as background templates. In [30], a *boundary connectivity* measure is introduced where a region's saliency is defined as being inversely related to the strength of its connectivity to the boundary of the image. In [31], image boundaries are also utilized as part of the object prior. In [29] a supervised approach is used where a set of features are integrated by learning to identify the discriminative ones using a Random Forest regressor. In this method too, the image borders are used to define a pseudo-background to calculate relevant discriptors. The method of [19] employs QCuts [17], a spectral clustering based object segmentation technique that hinges on Quantum Mechanical principles. In its application to salient object detection, a background prior based on image boundaries is used where the superpixels at the border of the image are assumed to belong to background. Together, all these methods either are supervised (relying on annotated training samples) or incorporate strong priors in their formulation that pertain only to natural images. Both these factors hinder their application to volumetric segmentation of digital porous media images as, unlike natural images, the region of interest (solids) is not limited to certain spatial locations in the lattice. However, QCuts is unique from the other methods in the sense that it is centered on a specialized graph-cut operation and lends itself to application on *any* graphical representation. Also, it is fully automatic and does not rely on labelled training examples for parameter selection or tuning. With the necessary reformulation to 3D, these two key factors obviously make the native QCuts the most promising method for the 3D pore segmentation. Therefore, the primary objective of this study was the development of a novel, highly accurate and fast graph-based extension, QCuts-3D, orchestrated exclusively for segmentation of digital porous media representations. The proposed method involves a single volume-to-volume mapping operation that is devoid of any parameter tuning, or redundant multiple-passes over the image. Furthermore, to alleviate the ambiguity surrounding the segmentation performance of different techniques, we introduce a vast collection of high-resolution 3D volumetric images of porous media, along with voxel-wise accurate ground truth annotation. This, for the first time, establishes a benchmark for porous media segmentation that enables precise quantitative comparisons among segmentation methods.

## 3 Proposed Methodology - QCuts-3D

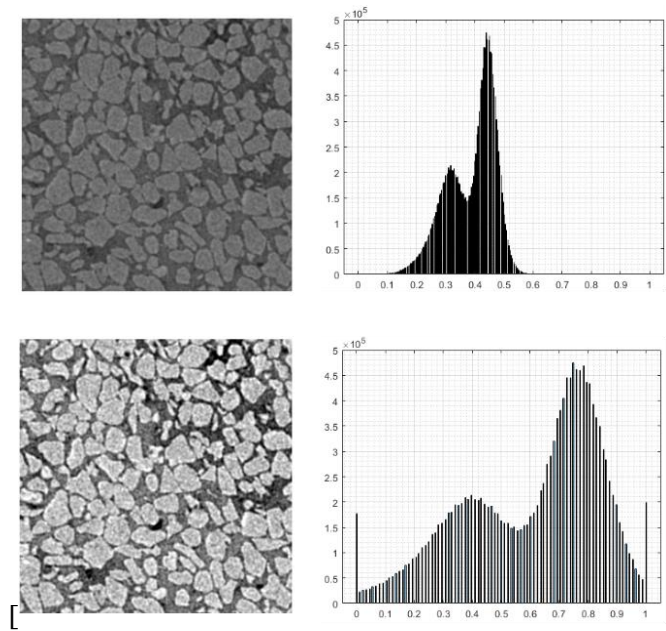

Figure 2 2D slice of a test original volume (top) and its contrast adjusted output (bottom) with the corresponding histograms (right).

In the proposed architecture, the contrast of the volumetric image is first adjusted as shown in Figure 2. Then QCuts-3D proceeds as follows: a supervoxel-based representation of the volumetric image is obtained at multiple scales. Then, for each scale, a graph is constructed where the nodes of the graph represent the supervoxels [34]. Using the graph cut technique of [18], a binary labeling of the nodes is then obtained, identifying them as either solid or pore. Results from each scale are finally combined in a voxel-wise majority voting scheme to obtain the desired segmentation. In the next section we shall briefly introduce the traditional QCuts method for salient-object segmentation. Then the following sections will introduce the extensions and modifications performed to accomplish QCuts-3D for the segmentation of 3D volumetric data.

## 3.1 Quantum Cuts

Image segmentation tasks are often posed as energy minimization problems where the aim is to find a labeling **y** that minimizes an energy function of the form:

$$E(\mathbf{y}) = E_{smooth}(\mathbf{y}) + E_{data}(\mathbf{y}) \qquad (1)$$

In (1), the data term $E_{data}(\mathbf{y})$ encourages the labeling to be consistent with the given image information while the smoothness term $E_{smooth}(\mathbf{y})$ ensures that the labeling is smooth over neighboring image elements [35]. In the widely used framework of active contours [26], the energy is a summation of an internal energy term (smoothness term), which controls the bending of the segmentation contour, and an external energy term (data term) based on local image features such as edges, which pushes the contour towards object boundaries. Despite their popularity, active contour-based methods are quite sensitive to parameters such as the weights in the energy function and the choice of initialization. Moreover, they do not guarantee a globally optimum solution and can only find the local minima closest to the initialized contour [36].

Graph cut-based techniques provide a globally optimum solution to many energy minimization problems related to image segmentation [37,38]. The energy function to be minimized takes the following general form, [39]:

$$E(\mathbf{y}) = E_{unary}(\mathbf{y}) + \lambda E_{binary}(\mathbf{y}) \qquad (2)$$

$$E_{unary}(\mathbf{y}) = \sum_i \phi(y_i) \qquad (3)$$

$$E_{binary}(\mathbf{y}) = \sum_{p,q} \psi(y_p, y_q) \qquad (4)$$

The unary (data) term $E_{unary}(\mathbf{y})$ measures how appropriate label $y_i$ is for the $i^{th}$ node given the local information about the image. The binary (smoothness) term $E_{binary}(y)$ measures the cost of cutting the edges, i.e., assigning disparate labels to connected neighbouring nodes $p$ and $q$, while $\lambda$ controls the weight of these terms. It can be noted that a binary term consisting of the summation of edges to be cut is not appropriate as it favours cutting short boundaries resulting in small isolated regions. To alleviate this, several domain-specific modifications to the energy function of (2) are generally made [40,41].

QCuts [17,19] is a graph cut method specifically tailored for discrete binary labeling vision problems. It produces state-of-the-art results for the task of salient-object segmentation where the aim is to identify the most visually appealing regions in an image. In QCuts, the modified energy function takes the form:

$$E_m = \frac{E_{unary}(\mathbf{y}) + E_{binary}(\mathbf{y})}{\sum_i y_i} \qquad (5)$$

In (5), the denominator term, $\sum_i y_i$, is introduced to maximize the area of the foreground/salient region. This is a key aspect of QCuts that makes it different from other widely used graph-cut based segmentation techniques such as [42] and [41]. Specifically, the fore-mentioned methods aim at partitioning the graph into two or more homogeonous regions where as QCuts' optimization criterion is more inclined towards separating nodes belonging to a single region of interest (foreground) from the rest [18]. Hence, it is more suitable for applications that can be posed as a one-vs-rest categorization problem such as salient object detection and binary segmentation of porous media images.

Given a graphical representation of the image where the edge weight connecting node $i$ and $j$ is denoted as $w_{ij}$, the problem can be expressed alternatively as follows:

$$\underset{\mathbf{y}}{argmin} \; \frac{\sum_i \phi(y_i) + \sum_{i,j} w_{i,j}(y_j - y_i y_j)}{\sum_i y_i} \qquad (6)$$

In order to facilitate the minimization, the labeling vector **y** is replaced by another vector $\mathbf{z} = \mathbf{y} \circ \mathbf{y}$ which can take values in $[-1, 0, 1]$. Furthermore, an additional phase term is introduced to penalize sign changes of **z**. Hence, the problem now becomes the following minimization:

$$\underset{\mathbf{z}}{argmin} \; \frac{\sum_i \phi(z_i^2) + \sum_{i,j} w_{i,j}(z_j^2 - z_i^2 z_j^2) + \sum_{i,j} w_{i,j}(z_i^2 z_j^2 - z_i z_j)}{\sum_i z_i^2} \qquad (7)$$

$$= \underset{\mathbf{z}}{argmin} \; \frac{\sum_i \phi(z_i^2) + \sum_{i,j} w_{i,j}(z_j^2 - z_i z_j)}{\sum_i z_i^2} \qquad (8)$$

$$= \arg\min_{\mathbf{z}} \frac{\mathbf{z}^T (H_m) \mathbf{z}}{\mathbf{z}^T \mathbf{z}} \tag{9}$$

In (9), the matrix $H_m$ is expressed as follows:

$$H_m = \begin{cases} \phi(i) + \sum_{k \in N_i} w_{ik}, & if \quad i = j \\ -w_{ij}, & if \quad i \in N_i \\ 0, & otherwise \end{cases} \tag{10}$$

where $N_i$ denotes the nodes in the neighbourhood of node $i$. For any non zero $\phi(i)$, $H_m$ is a positive definite matrix. Therefore, if the solution set of the above problem is relaxed such that $\mathbf{z} \in \mathbb{R}$, the minimization can be treated as a Rayleigh quotient problem. The optimal labeling vector can then be obtained by solving the eigenvalue problem of (11) and using the Hadamard product of any eigenvector of $H_m$ corresponding to the smallest eigenvalue (12).

$$H_m \mathbf{z}^* = E_m \mathbf{z}^* \tag{11}$$

$$\mathbf{y}^* = \mathbf{z}^* \circ \mathbf{z}^* \tag{12}$$

The obtained solution using QCuts has a theoretical correspondence with a quantum mechanical particle's location in space [18,43], hence the name "Quantum Cuts".

## 3.2 Supervoxel-based representation in QCuts-3D

Given a contrast-adjusted volumetric image, we proceed by obtaining its supervoxel representation. Supervoxels (and their 2D counterparts; superpixels) are clusters of image elements grouped together based on similarity in low-level image properties such as color cues and spatial information. They are designed to be compact and of nearly uniform size while also preserving regional boundaries [44]. Graphical image representations based on supervoxels are used in a variety of computer vision tasks [45], especially related to segmentation. By using a local statistic, such as the average grayscale intensity of the constituting voxels, to be representative, supervoxels can be treated as single entities to be labelled. This alleviates the need for separately labeling individual voxels, which, in addition to being computationally expensive, is also prone to producing noisy segmentation results. Supervoxel-based representations have been used in a variety of segmentation tasks dealing with volumetric tomography data [28,39,46,47]

In the proposed approach, we obtain supervoxel based oversegmentations at 4 different resolutions, in order to account for the varying scale at which particles can be found in the studied porous media sample. For generating supervoxels, we use the SLIC0 method which is a parameter-free version of the original method proposed in [34]. The resolutions employed are 2000, 4000, 6000 and 8000 supervoxels per volumetric image. This step is followed by constructing a graph where nodes represent supervoxels.

## 3.3 Graph Construction in QCuts-3D

In traditional QCuts, each node is connected to up to its fifth set of neighbors in the spatial domain, which helps in encoding contrast and textural information. Such an approach is applicable in case of natural images where the object of interest is geometrically compact and texturally unique. However, in case of volumetric images of porous media, this is not generally true as the supervoxels belonging to the region of interest (solid) do not generally exhibit unique textural characteristics and are not necessarily placed close to each other but are rather dispersed at various locations inside the volume. This can be visually illustrated by comparing the region of interest for a natural image to that in a 2D slice of an image of porous media (Figure 3).

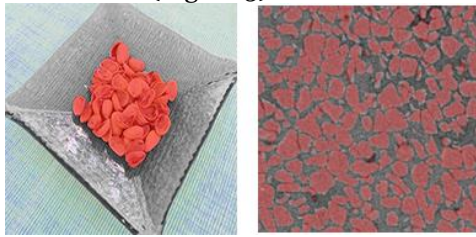

Figure 3 Comparison of the region of interest (highlighted in red) in a natural image (left) and a 2D slice from a test volumetric image of porous media.

In this context, QCuts-3D uses a fully connected graph where each node is connected to every other node in the graph. For defining the edge weights, we use the disparity in gray-scale intensities as the similarity metric. For each supervoxel, the average gray-scale intensity of all the constituting voxels is taken as its representative. Edge weights are then computed using a Gaussian kernel, as is commonly done in graph-based image processing [46,48]. The final edge weight assignment is mathematically expressed as follows:

$$w_{ij} = \exp -\frac{\left\|S_i - S_j\right\|_2}{2\sigma^2} \quad (13)$$

where $S_i$ is calculated as,

$$S_i = \frac{\sum \delta_{ik} I_k}{\sum \delta_{ik}} \quad (14)$$

In (14), $\delta_{ik}$ indicates whether $k_{th}$ voxel belongs to $i^{th}$ supervoxel

$$\delta_{ik} = \begin{pmatrix} 1, & if\ k^{th}\ voxel \in S_i. \\ 0, & \text{otherwise.} \end{pmatrix} \quad (15)$$

### 3.4 Unary Potentials in QCuts-3D

In (10), the unary term $\phi(i)$ encodes prior information about the labeling of nodes and is related to the potential of a node to belong to the background. In native QCuts, superpixels occupying the boundaries of the 2D image are assumed to belong to the background, based on the observation that objects of interest in natural images frequently tend to lie close to the center of images [16,49]. Therefore, $\phi(i)$ is set to a very high value for these nodes and zero for all the others. However, such a location-based prior is not suitable for the case of images of porous media. This is demonstrated in Figure 4 which compares a pixel-wise dataset-wide average of ground truth annotations of 3 popular salient object detection datasets and 2D slices of the SIMUPOR dataset. One can clearly observe that the salient object has a noticeable location bias as it tends to reside more towards the center and rarely at the borders of the image. However, for our task, the region-of-interest (solids) is spread across the entire frame and shows no bias for any particular spatial location.

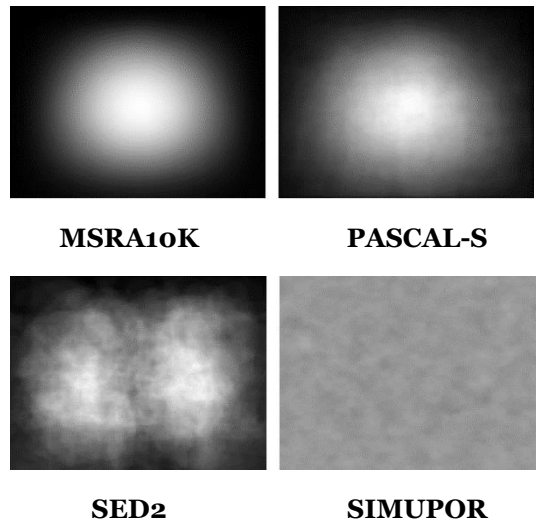

Figure 4 Mean ground truth annotations for [50],[51],[52] (reprinted from [16]) and SIMUPOR dataset.

Earlier, for such volumetric images, [53] proposed to use a manual selection of seeds for each phase by a skilled operator. However, the authors themselves identified a need for an automated process, as manual selection makes the overall process expensive in terms of processing time and also influences the segmentation results based on the operator's subjectivity. Moreover, owing to limitations in visualization, manual selection of seeds is not practical in the case of 3D images [54].

In light of the aforementioned facts, in QCuts-3D, the only assumption we make about the pore space is that it occupies the lower end of the grayscale intensity distribution. Exploiting this, we slice the volumetric image along the longitudinal axis and for each slice, we perform a row-wise selection of the supervoxel with the lowest mean intensity. This slice-wise approach provides a computationally economical and parameter independent way to make sure that the initial seed selection is not affected by the global intensity variations in the given tomography data.

### 3.5 Binarization in QCuts-3D

For each supervoxel resolution, we construct the graph, identify pore seeds, and then perform QCuts-3D. The initial output is a real-valued labeling vector which assigns a probability to each supervoxel based on its likelihood of belonging to solid space. In order to convert this to a binary labeling, we apply a binary k-means [55] clustering operation on the labeling vector. The final output of QCuts-3D is formed by labelling the supervoxels corresponding to nodes inside the cluster with the highest average probability as "solid" while the rest are labelled as "pores". Figure 5 illustrates the proposed end-to-end pipeline.

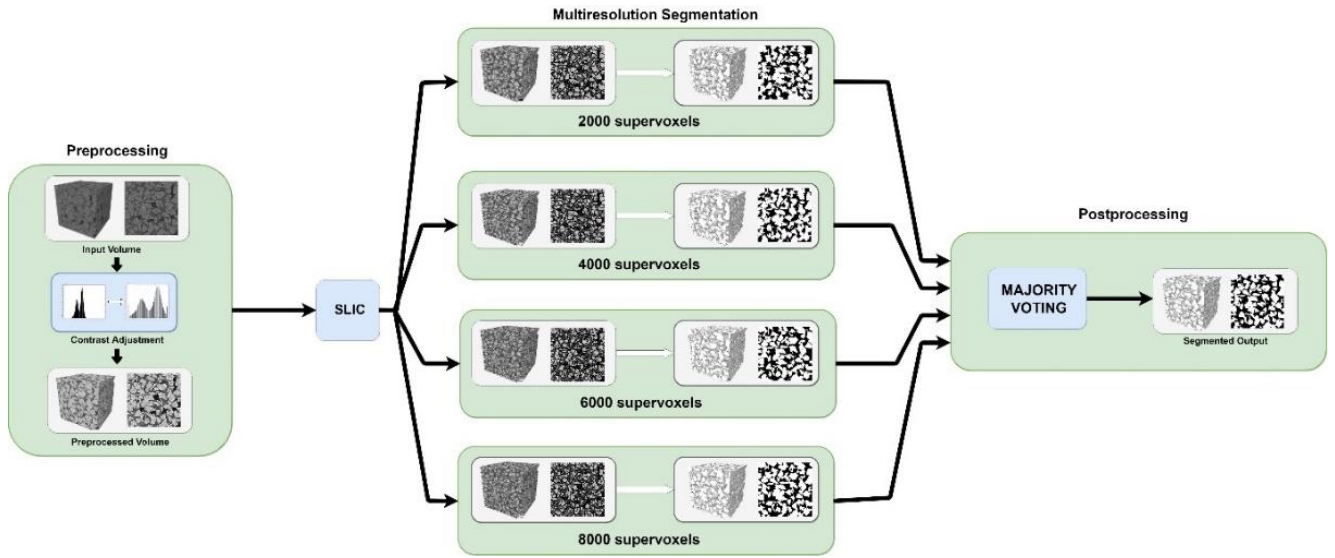

Figure 5 The illustration of the QCuts-3D for segmentation of volumetric images of porous media.

## 4    Experimental Results

### 4.1 The Benchmark Dataset: SIMUPOR

In this study, a benchmark dataset (aka SIMUPOR) of 68 3D volumetric images of porous media with varying grain geometry and composition is composed[1]. 3D volumetric images were obtained from the experiments conducted in [56] to study the effect of grain geometry on the morphology of non-aqueous phase liquids in porous media. The volumes correspond to samples from 34 different experiments, each corresponding to a specific constitution of porous medium. Among the 68 volumetric images used, 40 belong to the experiments that employ silica sand to model the porous media whereas the remaining 28 used quartz crystals. In addition to this variation in the shape of grains, there is also a variety in the size, with the median grain diameter ranging from 0.179 to 0.433 mm. This provides a comprehensive benchmark to check for robustness of any segmentation algorithms to changes in porous media composition.

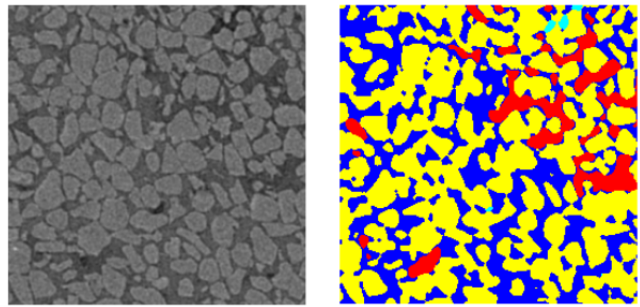

Figure 6  2D slice of a sample image from SIMUPOR dataset and its manually annotated multi-phase ground truth. Solids are color-labelled as yellow while the other porous phases (water, oil and gas) are labelled in blue, red and cyan color respectively.

---

[1] Dataset is publicly available at http://104.131.82.198/simupor/

A multi-phase ground truth segmentation for the volumetric images was obtained by applying a manual operator-guided process based on the indicator kriging approach [10]. For each phase, images obtained from the samples scanned at different energy levels were aligned and subtracted in order to emphasize that particular phase. Afterwards, segmentation for that phase was obtained by using the method of [10]. Furthermore, the accuracy of the segmented output was verified by comparing the porosity values for each phase. A visual example showing a slice of a test image along with its annotated ground truth is presented in Figure 6. For more details about the image acquisition and annotation process, the reader is referred to [56].

### 4.2 Evaluation Metrics

To perform the quantitative evaluation, we calculate a variety of evaluation metrics to precisely measure the performance of each method. A brief description of each metric used and their mathematical formulation are presented as follows:

*4.2.1 Jaccard Index*

As originally proposed in [57], the Jaccard Index is used as a similarity measure between two sets. In the context of Computer Vision, it is often used to compare the obtained segmentation results with the ground truth masks [58,59]. Also referred to as, "Intersection over Union" (IoU), it is mathematically formulated as follows:

$$J(\hat{y}, y) = \frac{|\hat{y} \cap y|}{|\hat{y} \cup y|} \quad (16)$$

In (15), $\hat{y}$ is the machine produced segmentation, $y$ refers to the ground truth mask and $|.|$ is the cardinality operator.

*4.2.2 Receiver Operating Characteristics (ROC)*

ROC curves are an important tool in visualizing classifier performance. It is a two-dimensional plot with true-positive rate (TPR) plotted on the Y-axis and the false positive rate (FPR) plotted on the x-axis.

Mathematical expressions for TPR and FPR are as follows:

$$TPR = \frac{True\ Positives}{True\ Positives + False\ Negatives} \quad (17)$$

$$FPR = \frac{False\ Positives}{True\ Negatives + False\ Positives} \quad (18)$$

The extremes (0,0) and (1,1) in the ROC plot correspond to cases where no positive classification is issued and where all classifications are positive, respectively. The upper left corner (0,1) represents the ideal case where there is no misclassification. An interesting property of the ROC curve is the area under it (AUROC). It corresponds to the probability of assigning a higher score to a randomly chosen positive instance as compared to a randomly chosen negative instance [59]. Being a scalar, the AUROC provides a convenient way to compare classifier performance.

*4.2.3 Misclassification Error (ME)*

Misclassification error is a simple measure often employed to measure the accuracy of the segmented output. It is given as the fraction of voxels that are classified incorrectly. The mathematical expression is as follows:

$$ME = \frac{False\ Positives + False\ Negatives}{Total\ number\ of\ voxels} \quad (19)$$

### 4.3 Performance Evaluations

Table 1 and Table 2 chronicle the results of our experiments over the SIMUPOR dataset by using the aforementioned evaluation metrics. Table 2 presents the performance of the state-of-the-art automatic segmentation method of [13] and QCuts-3D on images categorized with respect to different grain shapes. On the average, QCuts-3D achieves a boost of 26% in AUROC, 23.5% in IoU and 68.9% in ME. Similarly, Table 2 shows the same performance criteria evaluated across a variety of grain sizes. The proposed method achieves a boost of 27% in AUROC, 23% in IoU and a 70% decrease in ME, on the average. Furthermore, we also observe a decrease of 28.2%, 16.1% and 28.9% in the standard deviation of AUROC, IoU and ME, respectively. The results indicate a significantly superior performance in terms of accuracy and consistency. The latter is particularly important as QCuts-3D is more robust to the variations in the grain size as compared to the state-of-the-art method [13].

Table 1 Performance of QCuts-3D and the competing method [13] across different grain sizes.

|  | d1 | | d2 | | d3 | | d4 | | d5 | | d6 | |
|---|---|---|---|---|---|---|---|---|---|---|---|---|
|  | [13] | QCuts-3D | [13] | QCuts-3D | [13] | QCuts-3D | [13] | QCuts-3D | [13] | QCuts-3D | [13] | QCuts-3D |
| AUROC | 0.7327 | 0.9430 | 0.7222 | 0.9138 | 0.7097 | 0.9282 | 0.7544 | 0.9387 | 0.7226 | 0.9383 | 0.7437 | 0.9200 |
| IoU | 0.7310 | 0.9147 | 0.7095 | 0.8781 | 0.7308 | 0.9009 | 0.7568 | 0.9087 | 0.7579 | 0.9210 | 0.7000 | 0.8722 |
| ME | 0.2180 | 0.0527 | 0.2363 | 0.0778 | 0.2270 | 0.0647 | 0.1944 | 0.0572 | 0.2058 | 0.0530 | 0.2318 | 0.0772 |

In terms of qualitative comparison, Figure 7 presents the results of [13] and QCuts-3D on the images from the SIMUPOR dataset. One can clearly observe that, owing to the presence of multiple phases in the images, the adaptive thresholding technique of [13] fails to provide adequate results, which is not surprising at all, given the fact that this method assumes *a priori* the existence of only two distinct populations of voxels in an image. The method fares well in identifying the gas as the pore phase, but falls short when it comes to liquid or oil phases, which have intensities closer to the solid phase.

Table 2 Performance of QCuts-3D and the competing method [13] across different grain sizes.

|  | Sands | | Quartz | | Average | |
|---|---|---|---|---|---|---|
|  | [13] | QCuts-3D | [13] | QCuts-3D | [13] | QCuts-3D |
| AUROC | 0.7185 | **0.9209** | 0.7400 | **0.9250** | 0.7293 | **0.9229** |
| IoU | 0.7254 | **0.8917** | 0.7131 | **0.8850** | 0.7192 | **0.8883** |
| ME | 0.2273 | **0.0705** | 0.2271 | **0.0706** | 0.2272 | **0.0705** |

Furthermore, the grain boundaries are also not delineated clearly as compared to the QCuts-3D. This can be attributed to the fact that local thresholding within cubical windows does not take into account regional characteristics whereas the proposed supervoxel-based representation already incorporates edge preservation and thus ensures that the regional intricacies are respected in the final segmentation outcome.

### 4.4 Computational Complexity Analysis

MATLAB implementation of QCuts-3D and [13] were tested in MATLAB R2018a on an Intel Xeon CPU E5-2690 processor running at 2.60 GHz with 64 GB of memory. Computational efficiency for both methods was measured by applying them over the entire SIMUPOR dataset and averaging the overall run-times. We observe that, on the average, the competing method [13] took 937.85 minutes on a single volumetric image of size 256x256x256 whereas the proposed multi-resolution scheme took only 100.96 seconds. This presents around 500-fold less computational time due to the fact that the competing method involves multiple passes over the entire image as part of its search for optimal window size, which has a substantial increase on the computational complexity. In contrast, QCuts-3D is devised from the ground up to consist of efficient end-to-end graph-cut operations and is kept free from any need to repeatedly revisit the image space for parameter optimization.

## 5 Graph Fourier Transform Perspective and Multi-Phase Segmentation

The correspondence of QCuts with spectral clustering techniques was first noted in the original work [24]. Specifically, the matrix $H_m$ in (10) can be written as:

$$H_m = D - W + V \qquad (20)$$

In (20), D is a square matrix with size equal to the number of nodes in the graph, whose each diagonal entry $d_{ii}$ is the summation of edge weights incident on node $i$. W is the weight matrix whose elements $w_{ij}$ correspond to the edge weight connecting node $i$ and $j$, while $V$ is a diagonal matrix encoding the unary potentials $\phi(i)$ as its diagonal entries. A widely used matrix describing the graph is the Laplacian which is denoted as $L = D - W$. $H_m$ can then be considered as a modified form of Laplacian with an additional term $V$ to incorporate prior potentials. In the recent literature, the notion of graph Fourier transform (GFT) is introduced where the eigenvectors of the graph Laplacian are considered as the graph Fourier basis functions with the corresponding eigenvalues providing a notion of graph frequencies [60,61]. A low-rank approximation of smooth graph signals (signals defined over the graph) using these basis functions lends itself to useful applications in digital signal processing [60,62,63].

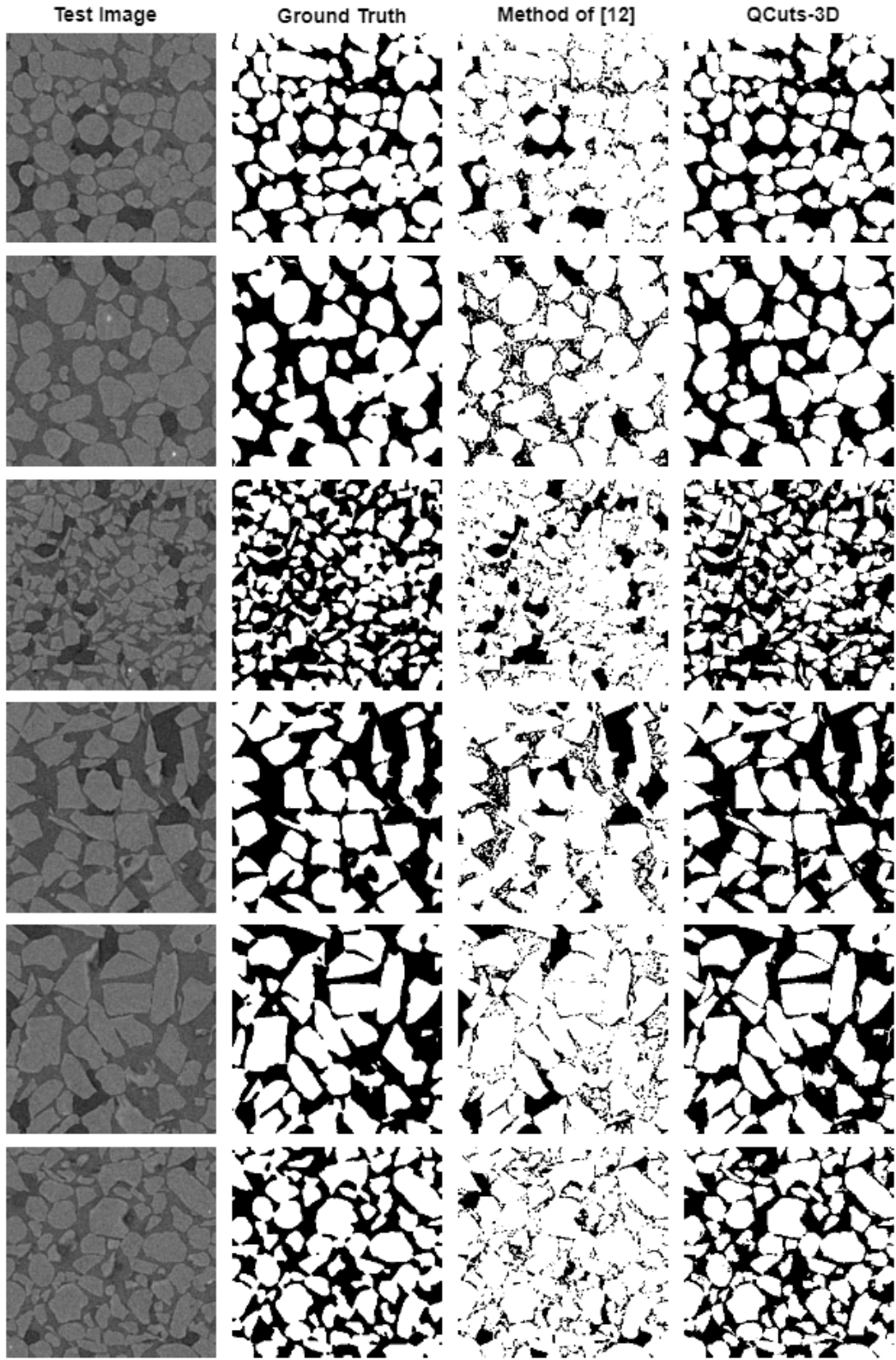

Figure 7 Compilation of 2D slices of test images (rows) from the SIMUPOR dataset, ground truth, segmentation output of [12] and that of QCuts-3D (from left to right).

Given a graphical representation of a 2D image similar to the one proposed in [18], the authors of [64] linearly superposed multiple low-frequency graph Fourier basis functions to generate category-independent object proposals. In [65], saliency maps obtained from existing salient-object segmentation methods were denoised by projecting them onto the space spanned by the low-frequency basis functions. The results showed that most of the useful information pertaining to the region of interest is encoded in the lower part of the frequency spectrum. Extending the same principle to multiphase volumetric images, we can claim that an accurate segmentation for any region of interest (phase) can be obtained by combining the low-frequency basis functions. Using the graph construction as described in Section 3.3, we test this claim by projecting the supervoxel representation of ground truth segmentation for each phase onto the space spanned by a progressively higher number of low-frequency basis functions, as proposed in [65]. Figure 8 shows the mean squared error of this reconstructed output as compared to the original ground truth for each phase against varying degrees of spectrum utilized.

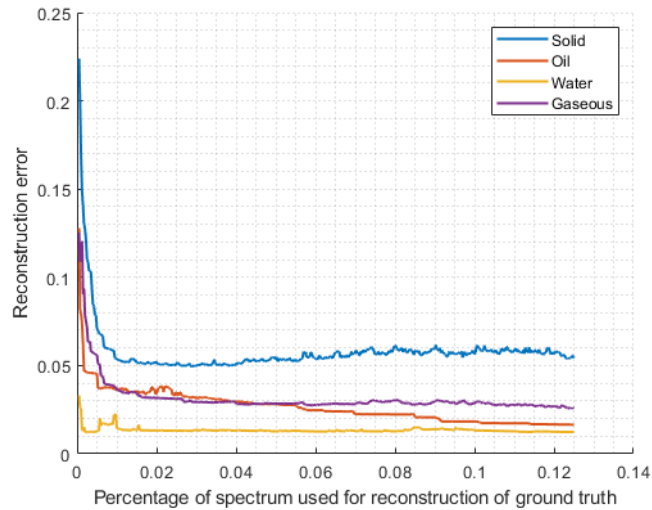

Figure 8 Average reconstruction error for each phase over the SIMUPOR dataset plotted against the percentage of spectrum utilized.

We see that the ground truth for each phase can be accurately approximated by using only a few basis functions from the lower frequency band. This provides an interesting future direction where, using the ground truth, we can learn to combine the low-frequency basis functions in a supervised manner to produce more accurate multi-phase segmentations. This will be the topic of our future work.

## 6  Conclusions and Future Work

In this work, we proposed a novel, automatic and unsupervised 3D segmentation technique for digital volumetric images of porous media. We also presented a benchmark dataset consisting of 68 such volumetric images with corresponding voxel-wise segmentation ground truths for multiple phases. We performed an extensive comparative evaluation over this dataset of the proposed QCuts-3D against the state-of-the-art segmentation method [13]. Both quantitave and qualitative results revealed that QCuts-3D achieves a significant improvement in segmentation accuracy and in computational efficiency. Moreover, the proposed technique was also found to be more robust to variations in compositional elements of the pores and the geometry of grains composing the porous media. Visual comparison further revealed the fact that the regional boundaries are better preserved using the proposed supervoxel-based approach, as compared to localized thresholding in cubical windows employed by the competing method. Finally, the graphical representation of the volumetric image lends itself to an extension towards multiphase segmentation in a supervised approach.

Future work involves investigating ways of improving the computational efficiency of the proposed method and extending it towards application in multiphase segmentation. Currently, the primary computational bottleneck in the proposed pipeline is the multiresolution supervoxel-based oversegmentation. We aim to harness the power of GPUs to expedite and parallelize this step. Furthermore, we will also explore the possibility of devising new single-pass multiphase segmentation approaches by utilizing volumetric stacks of images acquired using scanning at different energy levels. Moreover, the introduction of the first benchmark (SIMUPOR) dataset can potentially usher in an era of rapid influx of modern segmentation techniques such as Deep Learning-based supervised approaches, that can learn more efficient graphical representations [66] to segment the images using the provided ground truth annotations.

## Acknowledgment

This publication was made possible by NPRP grant # NPRP9-390-1-088 from the Qatar national research fund (a member of Qatar Foundation). The findings achieved herein are solely the responsibility of the authors.

MALIK ET AL.: 3D QUANTUM CUTS FOR AUTOMATIC SEGMENTATION OF POROUS MEDIA IN TOMOGRAPHY IMAGES 16